\documentclass[runningheads]{llncs}

\usepackage[year=2026]{accv}
\usepackage{accvabbrv}
\usepackage{graphicx}
\usepackage{booktabs}
\usepackage{amsmath,amssymb}
\usepackage{multirow}
\usepackage{xcolor}
\usepackage[accsupp]{axessibility}
\usepackage{tikz}
\usetikzlibrary{positioning,fit,arrows.meta,shapes.multipart}
\usepackage[pagebackref,breaklinks,colorlinks,allcolors=blue]{hyperref}

\newcommand{\ours}{FastSlow-LMDrive}

\begin{document}

\title{Think at 5\,Hz, Act at 20\,Hz: Asynchronous Fast-Slow Vision-Language-Action Inference for Closed-Loop Driving}
\titlerunning{Asynchronous Fast-Slow VLA Inference for Closed-Loop Driving}

\author{Yun Li\inst{1} \and
Jiachen Gong\inst{1} \and
Simon Thompson\inst{2} \and
Ehsan Javanmardi\inst{1} \and
Qunli Zhang\inst{3} \and
Zifan Zeng\inst{3} \and
Shiming Liu\inst{3} \and
Peng Wang\inst{3} \and
Zixuan Guo\inst{1} \and
Manabu Tsukada\inst{1}}

\authorrunning{Y.~Li et al.}
\institute{The University of Tokyo, Tokyo, Japan\\
\email{li-yun@g.ecc.u-tokyo.ac.jp} \and
TIER IV, Inc., Tokyo, Japan \and
Huawei}

\maketitle

\begin{abstract}
Large language models bring instruction following and scene reasoning to
end-to-end driving, but their inference latency collides with the control
rate a vehicle requires. Existing closed-loop agents hide this gap by
invoking the model on alternate simulation ticks and replaying the previous
command in between, so half of all control outputs ignore the newest
observations. We present a fast-slow architecture that removes this
compromise. A frozen 7B vision-language backbone acts as the slow system,
digesting navigation instructions and visual history at low frequency while
exposing its per-layer key-value cache as a standing representation of the
scene. A lightweight action expert acts as the fast system, attending to
this cache and to the current camera frame at every simulation tick to
regress waypoints in a single forward pass. Since the cache lags behind the
world at deployment, we train the expert under randomized staleness,
aligning training with asynchronous execution. On LangAuto-Short routes in
CARLA, our system produces fresh control at every 50\,ms simulation tick
and lifts route completion from 37.0 to 94.0 over the frame-skipping
baseline. A frame-skip ablation with the same expert separates the two
factors at work: the expert raises the driving score on its own, while
per-tick freshness raises completion from 82.1 to 94.0 and cuts red-light
violations by a third. Trained on a single town, the
expert transfers zero-shot to two unseen towns, holding 84--94\% route
completion where the baseline reaches 31--41\%. It reduces open-loop
waypoint error by nearly a factor of four compared to the backbone's own
action head, at a per-tick model cost of 32\,ms that is independent of
history length on a single consumer GPU.
\keywords{Vision-language-action models \and Autonomous driving \and Real-time inference}
\end{abstract}

\section{Introduction}
\label{sec:intro}

\begin{figure}[t]
\centering
\includegraphics[width=\textwidth]{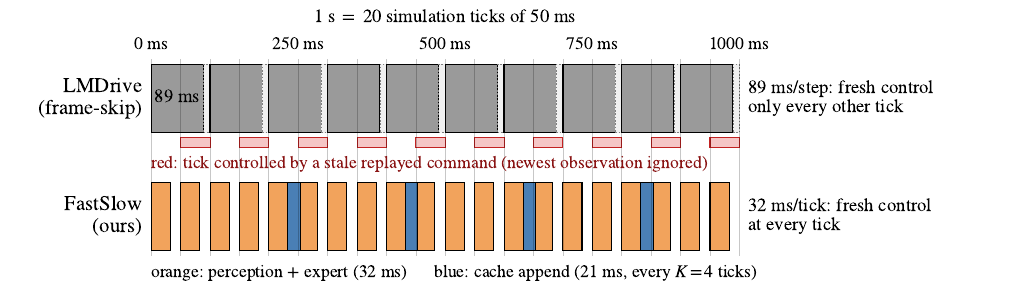}
\caption{One second of control under the two schedules, at the measured
per-stage latencies of Table~\ref{tab:latency}. The frame-skipping agent
spends 89\,ms per invocation, so it produces fresh control only at every
other 50\,ms tick and replays a stale command in between. Our expert reads
the backbone's cached context and issues fresh control at every tick for
32\,ms, plus a 21\,ms cache append every $K{=}4$ ticks.}
\label{fig:teaser}
\end{figure}

Language-conditioned driving agents built on large vision-language models
follow natural-language navigation instructions, explain their decisions,
and inherit world knowledge from internet-scale pretraining
\cite{shao2023lmdrive,xu2024drivegpt4,sima2023drivelm}. The price of this
capability is inference cost. A 7B backbone that re-reads its full visual
history at every control step cannot keep up with the 20\,Hz tick rate of a
closed-loop simulator, let alone a real vehicle. The canonical open-source
agent of this family, LMDrive \cite{shao2023lmdrive}, resolves the conflict
by running its model on every second tick and replaying the stale
control command in between. Half of the time, the vehicle acts on a
decision that ignores the most recent 50--100\,ms of observation.

\begin{figure}[t]
\centering
\includegraphics[width=\textwidth]{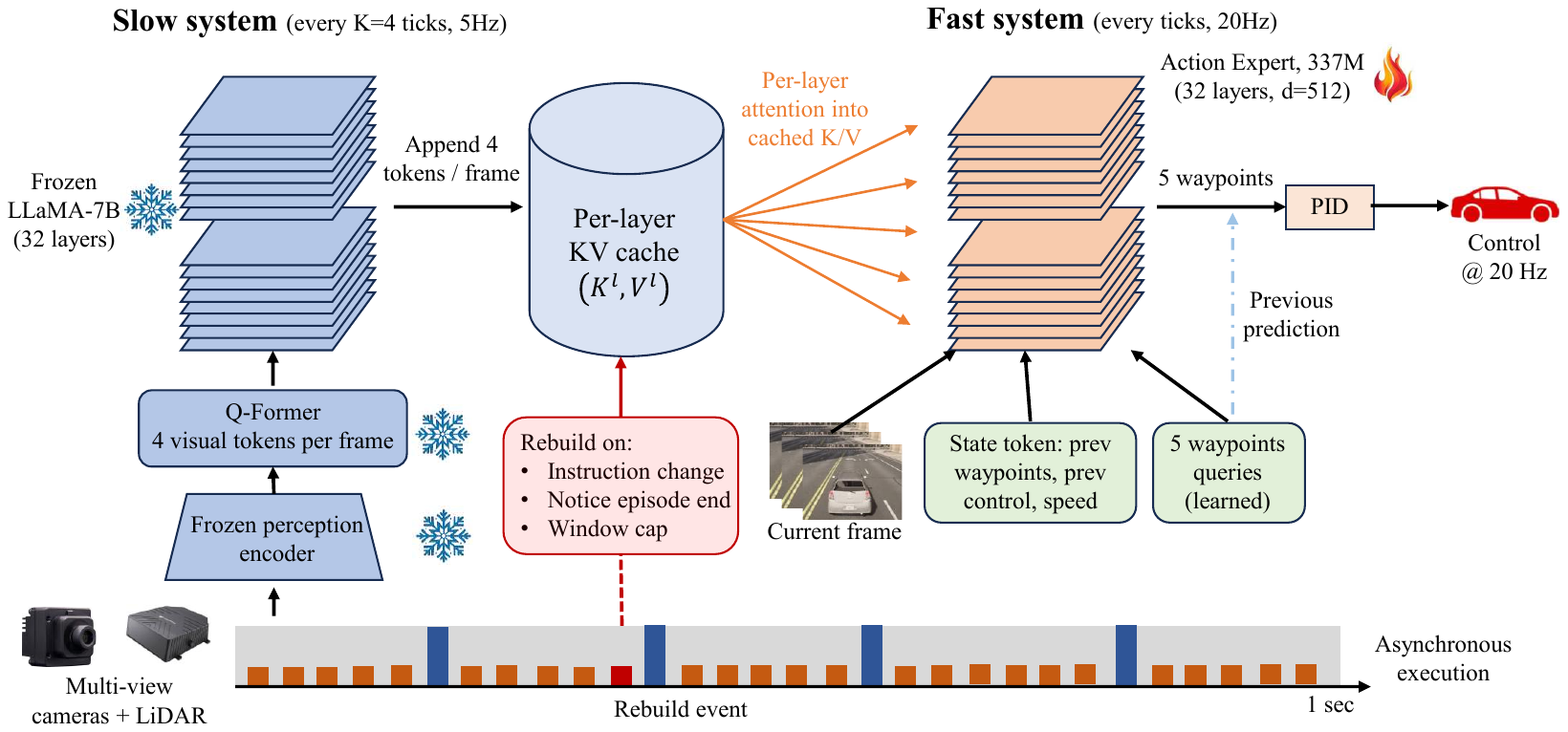}
\caption{Asynchronous fast-slow inference. The frozen backbone (left)
digests instruction and history every $K$ ticks, appending four tokens per
frame to its per-layer KV cache through an incremental forward pass; the
cache is rebuilt only on invalidation events. The action expert (right)
runs at every simulation tick: its ten tokens cross-attend into the cached
keys and values at each of the 32 layers and regress five waypoints. The
backbone never attends to expert tokens, so the cache is identical with or
without the expert.}
\label{fig:arch}
\end{figure}

We argue that the latency problem is architectural, not fundamental.
Instruction understanding and temporal scene aggregation, the expensive
parts of a VLA agent, change slowly, while mapping the current frame to a
trajectory must happen quickly yet is computationally light.
This observation suggests a fast-slow decomposition
(Figure~\ref{fig:arch}) in the spirit of
dual-process designs explored for driving by DriveVLM \cite{tian2024drivevlm}
and for manipulation by $\pi_0$ \cite{black2024pi0}.

 Recent work on action
experts, notably DriveVLA-W0 \cite{li2026drivevlaw0}, pairs a large VLA
backbone with a small decoder through joint attention, but evaluates only
open-loop planning benchmarks where all computation is synchronous and
latency is measured rather than lived. Whether a fast-slow VLA can absorb
cache staleness and still improve driving outcomes at full control rate on
commodity hardware has not been demonstrated.

This paper provides that demonstration. We attach a 337M-parameter action
expert to a frozen LMDrive backbone. Every layer of the expert projects its
tokens into the backbone's attention space and cross-attends into the
backbone's per-layer key-value cache; the backbone never attends back, so
the cache evolves exactly as it would alone. This asymmetry enables
asynchronous deployment. The slow system appends one frame to its cache
every $K$ ticks as an incremental four-token forward pass, rebuilding the
cache only when the instruction, a runtime notice, or an episode boundary
invalidates it, while the fast system reads whatever cache exists at every
tick. Because the cache the expert reads lags the world by up to $K$ ticks,
we train the expert under randomized staleness, truncating the backbone
prefix visible to each supervised frame by a random number of frames so the
expert must compensate through the current frame and an ego-state token
that carries its own previous prediction. A golden equivalence test
confirms that incremental caching is numerically faithful, since switching
from monolithic to incremental prefill moves predicted waypoints by less
than 4\,mm.

On the LangAuto-Short benchmark in CARLA \cite{dosovitskiy2017carla}, our
agent issues fresh control at every simulation tick on a single
RTX~3090~Ti and lifts the driving score from 28.8 to 32.9 while raising
route completion from 37.0 to 94.0 (two runs per configuration). A
frame-skip ablation running the same expert at the baseline's 10\,Hz
cadence attributes the driving-score gain to the expert itself, and the
completion gain, the near-elimination of route deviations and timeouts,
and a third fewer red-light violations to per-tick freshness. Per-tick
operation is itself available only through the cache: any head attached
to a backbone that recomputes its history spends 89--169\,ms per step and
must skip frames to fit the budget. In open
loop, the expert reduces
validation waypoint L1 error from 0.123 to 0.031, nearly a factor of four
below the frozen backbone's own action head, with only 337M trained
parameters and five epochs of training on a single GPU. Ablations isolate the contribution of staleness
augmentation, which improves validation waypoint error at every training
epoch and ends at 0.031 against 0.037 for a $\delta{=}0$-only twin.

We make three contributions. We introduce an asynchronous fast-slow
inference scheme for VLA driving that turns the backbone into a cached,
incrementally updated context provider and moves all per-tick computation
into a lightweight expert. We propose staleness-augmented training, which
aligns the training distribution with the asynchronous deployment
distribution. We present the first closed-loop evaluation of the fast-slow
VLA paradigm under measured latency on consumer hardware.

\section{Related Work}
\label{sec:related}

\subsubsection{Language-guided end-to-end driving.}
Sensor-fusion policies that operate without any language interface,
including TransFuser~\cite{prakash2021multimodal,chitta2023transfuser},
InterFuser~\cite{shao2022interfuser}, TCP~\cite{wu2022tcp}, and
ReasonNet~\cite{shao2023reasonnet}, remain strong closed-loop baselines on
CARLA, so a language-guided system must justify the latency of its
reasoning stack. Coupling driving policies to language models began with
interpretability ---
DriveGPT4 \cite{xu2024drivegpt4} narrates and predicts controls, DriveLM
\cite{sima2023drivelm} structures scene understanding as graph question
answering --- and matured into closed-loop instruction following with
LMDrive \cite{shao2023lmdrive}, which drives in CARLA from natural-language
commands and multi-modal sensor streams. GPT-Driver \cite{mao2023gptdriver}
recasts motion planning as language modeling, Agent-Driver
\cite{mao2024agentdriver} surrounds the planner with tool use and memory,
and EMMA \cite{hwang2024emma} maps camera input directly to trajectories,
but these are evaluated open-loop on recorded logs where inference latency
carries no penalty. Vision-only successors such as CarLLaVA
\cite{renz2024carllava} and SimLingo \cite{renz2025simlingo} raised
closed-loop benchmark scores while keeping the full language model inside
the control loop. None of these systems addresses the resulting rate mismatch;
in practice their public agents throttle the simulator or replay stale
commands between model invocations. Our work keeps the LMDrive stack intact
and changes only where computation happens in time.

\subsubsection{Fast-slow architectures and action experts.}
Dual-process designs separate deliberate reasoning from fast reaction.
DriveVLM \cite{tian2024drivevlm} pairs a VLM planner with a classical
trajectory refiner, but the two subsystems exchange only trajectories; the
fast path never sees the slow system's internal representation. In
manipulation, $\pi_0$ \cite{black2024pi0} couples a VLM with a flow-matching
action expert through joint attention inside one synchronous forward pass,
and action-chunking policies such as ACT \cite{zhao2023learning}, Diffusion
Policy \cite{chi2023diffusion}, and $\pi_{0.5}$ \cite{intelligence2025pi05}
decouple decision rate from control rate by emitting several future actions
per inference call.
DriveVLA-W0 \cite{li2026drivevlaw0} brings this coupling to driving and
shows that an asymmetric variant in which the backbone never attends to
expert tokens preserves planning quality on NAVSIM, yet both training and
evaluation remain synchronous and open-loop. We take the asymmetry to its
logical conclusion. Since the backbone's computation is independent of the
expert, its key-value cache can be reused across many expert invocations,
which turns the synchronous coupling into an asynchronous fast-slow system
with the backbone's per-layer cache as the interface. The expert must now
tolerate stale context, which we address with randomized staleness, and
the evaluation can close the loop under measured latency on consumer
hardware.

\subsubsection{Efficient inference for control.}
Key-value caching and incremental decoding are standard machinery for
autoregressive text generation, and robotics has begun to exploit them for
real-time action chunking \cite{black2024pi0}. Acting on inputs that lag
the world is a classical concern in control, studied as Markov decision
processes with observation and action delays
\cite{katsikopoulos2003markov}, and reinforcement learning has revisited
the setting by letting the environment evolve while the agent computes
\cite{ramstedt2019realtime,xiao2020thinking}. Staleness-augmented training
is the imitation-learning counterpart of these ideas for a cached context. Orthogonal strategies such
as distillation, quantization, and speculative decoding shrink or
parallelize the model itself. We instead reorganize when each component
runs, leaving the backbone neither compressed nor retrained and moving the
entire per-tick budget into a 337M expert whose cost is independent of
history length.

\section{Method}
\label{sec:method}

Figure~\ref{fig:arch} gives an overview. We build on LMDrive
\cite{shao2023lmdrive}. A frozen perception encoder converts multi-view
camera images and LiDAR into per-frame features, a Q-Former
\cite{li2023blip2} compresses each frame into $n{=}4$ visual tokens
conditioned on the instruction, and a frozen LLaMA-7B backbone
\cite{touvron2023llama} consumes the token sequence
$[\,\mathrm{text} \mid f_1 \mid f_2 \mid \cdots\,]$, optionally with a
runtime notice inserted mid-sequence. The original agent regresses waypoints
from the backbone's final hidden states with a small MLP head and recomputes
the entire sequence at every invocation.

\subsection{Action Expert with Per-Layer Cache Attention}
\label{sec:expert}

The expert is a 32-layer transformer with hidden width
$d_{\mathrm{ae}}{=}512$, matched one-to-one with backbone layers. At layer
$\ell$ it forms queries, keys, and values from its own tokens through
rectangular projections $W^{\ell}_{q,k,v}\in\mathbb{R}^{4096\times512}$ into
the backbone's 32-head, 128-dimensional attention geometry, concatenates the
backbone's cached keys and values of the same layer, and performs a single
attention operation
\begin{equation}
\mathrm{Attn}\!\left(Q_{\mathrm{ae}}^{\ell},\;
[K_{\mathrm{vlm}}^{\ell};K_{\mathrm{ae}}^{\ell}],\;
[V_{\mathrm{vlm}}^{\ell};V_{\mathrm{ae}}^{\ell}]\right),
\end{equation}
followed by an output projection back to $d_{\mathrm{ae}}$ and a SwiGLU
feed-forward block. The backbone never attends to expert tokens, so the
cached $K_{\mathrm{vlm}},V_{\mathrm{vlm}}$ are identical whether or not the
expert runs, the property that permits caching. Expert tokens carry
learned positional embeddings and no rotary encoding, following the design
validated by DriveVLA-W0 \cite{li2026drivevlaw0}; backbone keys retain the
rotary phases they were cached with.

Per frame the expert consumes ten tokens: one state token embedding the
previous predicted waypoints, the previous control triplet, current speed,
and the target point; the current frame's four visual tokens, produced by
the same Q-Former pipeline the backbone uses; and five learnable waypoint
queries, decoded to five $(x,y)$ waypoints by a two-layer head under L1
supervision. The expert totals 337M parameters. The backbone, Q-Former, and
original heads stay frozen, so training builds no autograd graph through
the 7B model and peaks below 28\,GB of GPU memory at batch size four.

\subsection{Staleness-Augmented Training}
\label{sec:staleness}

At deployment the cache lags the current frame by up to $K$ ticks. We
replicate this gap during training by drawing
$\delta\sim\mathcal{U}\{0,\dots,\delta_{\max}\}$ for each sample and
masking the backbone prefix visible to frame $j$'s expert block so that it
ends at frame $j-\delta$, or at the instruction text alone when
$j<\delta$. The expert
therefore learns to fuse an outdated scene summary with fresh per-tick
evidence. The state token is trained with teacher-forced previous waypoints
under Gaussian noise and random dropout, limiting exposure bias. All frames
of a training clip are supervised in parallel through per-block attention
masks, which reproduces the cache visibility of per-frame deployment while
leaving the data pipeline untouched, since the previous-action fields the state token needs
already exist in LMDrive's driving logs.

\subsection{Asynchronous Deployment}
\label{sec:deploy}

The agent maintains the backbone cache as mutable state. Every $K{=}4$
ticks (0.2\,s at 20\,Hz, matching the training frame grid) the current
frame's four tokens are appended through an incremental forward pass whose
cost is independent of history length; the episode-boundary classifier runs
on the resulting hidden states. The cache is rebuilt from scratch only on
instruction change, notice arrival, episode end, or when the cached history
reaches the training window cap. At every tick the expert reads the cache
as-is, so test-time staleness is bounded by $K{-}1$ ticks and stays inside
the training distribution. Rotary phases make append-only caching exact. A
golden test comparing monolithic against incremental prefill confirms
this, with waypoint differences below 4\,mm at the fp16 noise floor.

\begin{figure}[t]
\centering
\includegraphics[width=\linewidth]{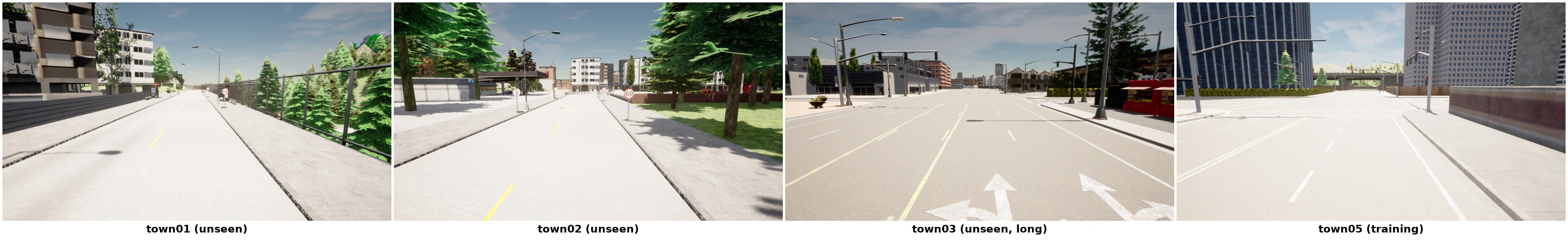}
\caption{The four evaluation towns. The expert is trained on short routes
from town05 only; towns 01, 02, and 03 are never seen during expert
training, and town03 is evaluated on the long-route tier.}
\label{fig:towns}
\end{figure}

\section{Experiments}
\label{sec:experiments}

\subsection{Setup}
\label{sec:setup}
We train on the town05 portion of the LMDrive corpus: 27{,}485 instruction
clips of up to 40 frames, with three weather conditions held out for
validation. The expert trains for five epochs on one 48\,GB GPU with AdamW,
a cosine schedule from $10^{-4}$, and an effective batch size of 32; the
backbone, perception encoder, Q-Former, and original heads stay frozen
throughout, initialized from the public LMDrive checkpoint. Closed-loop
evaluation uses the 32 LangAuto-Short town05 routes in CARLA 0.9.10 on a
single RTX~3090~Ti, with identical simulator settings, scenario
definitions, and PID parameters for every agent; the baseline is the public
LMDrive agent, which invokes its model on alternate ticks and replays the
previous command otherwise. We report the standard driving score, route
completion, and infraction score, plus per-stage latency measured both in a
controlled sweep over history lengths and inside the running loop.

\subsection{Open-Loop Waypoint Accuracy}
\begin{table}[t]
\centering
\caption{Validation waypoint L1 (m) on held-out weathers, all evaluated
under the synchronous $\delta{=}0$ protocol with teacher-forced state
tokens.}
\label{tab:openloop}
\begin{tabular}{lcc}
\toprule
Method & L1 $\downarrow$ & Params trained \\
\midrule
Backbone action head (frozen) & 0.123 & --- \\
Action expert, $\delta{=}0$ only & 0.037 & 337M \\
Action expert, randomized $\delta$ (ours) & \textbf{0.031} & 337M \\
\bottomrule
\end{tabular}
\end{table}

All rows share the synchronous $\delta{=}0$ protocol, in which the state
token receives teacher-forced previous waypoints, an input the frozen
backbone head does not consume; part of the open-loop gap therefore
reflects this privileged signal. The closed-loop comparison below, where
the expert sees only its own predictions, confirms the gain without
teacher forcing.

\subsection{Closed-Loop Driving}
\begin{table}[t]
\centering
\caption{LangAuto-Short, town05, 32 routes, identical simulator settings.
DS: driving score; RC: route completion; IS: infraction score. Baseline
and ours are means $\pm$ half-range over two independent runs each; the
frame-skipped middle row runs the same trained expert at the baseline's
10\,Hz replay cadence (single run) and isolates control freshness from
expert quality.}
\label{tab:closedloop}
\begin{tabular}{lcccc}
\toprule
Agent & Control rate & DS $\uparrow$ & RC $\uparrow$ & IS $\uparrow$ \\
\midrule
LMDrive (replayed ticks) & 10\,Hz & 28.8\,$\pm$\,0.8 & 37.0\,$\pm$\,0.4 & \textbf{0.80}\,$\pm$\,0.04 \\
\ours{}, frame-skipped & 10\,Hz & \textbf{34.0} & 82.1 & 0.45 \\
\ours{} (ours) & 20\,Hz & 32.9\,$\pm$\,0.7 & \textbf{94.0}\,$\pm$\,2.6 & 0.37\,$\pm$\,0.02 \\
\bottomrule
\end{tabular}
\end{table}

The completion gap dominates the comparison. The baseline accumulates 25.0
route deviations and 2.9 timeouts per kilometer, symptoms of acting on
stale decisions, against 4.3 and 0.08 for ours (two-run means). Our
infraction score is lower, and exposure explains only part of it. A
vehicle that completes 2.5 times more route meets traffic the baseline
never reaches, but distance normalization does not remove the gap, since
per-kilometer vehicle collisions still rise from 3.2 to 11.2; the expert
completes routes through traffic it has not learned to negotiate safely,
a trade-off we revisit in the transfer experiments and the limitations.
Infractions tied to inattention move the other way, with layout
collisions dropping from 10.3 to 2.2 per kilometer and off-lane driving
from 7.9 to 1.1. The low-level PID controller is reused from the 10\,Hz
baseline without retuning.

\begin{figure}[t]
\centering
\includegraphics[width=\linewidth]{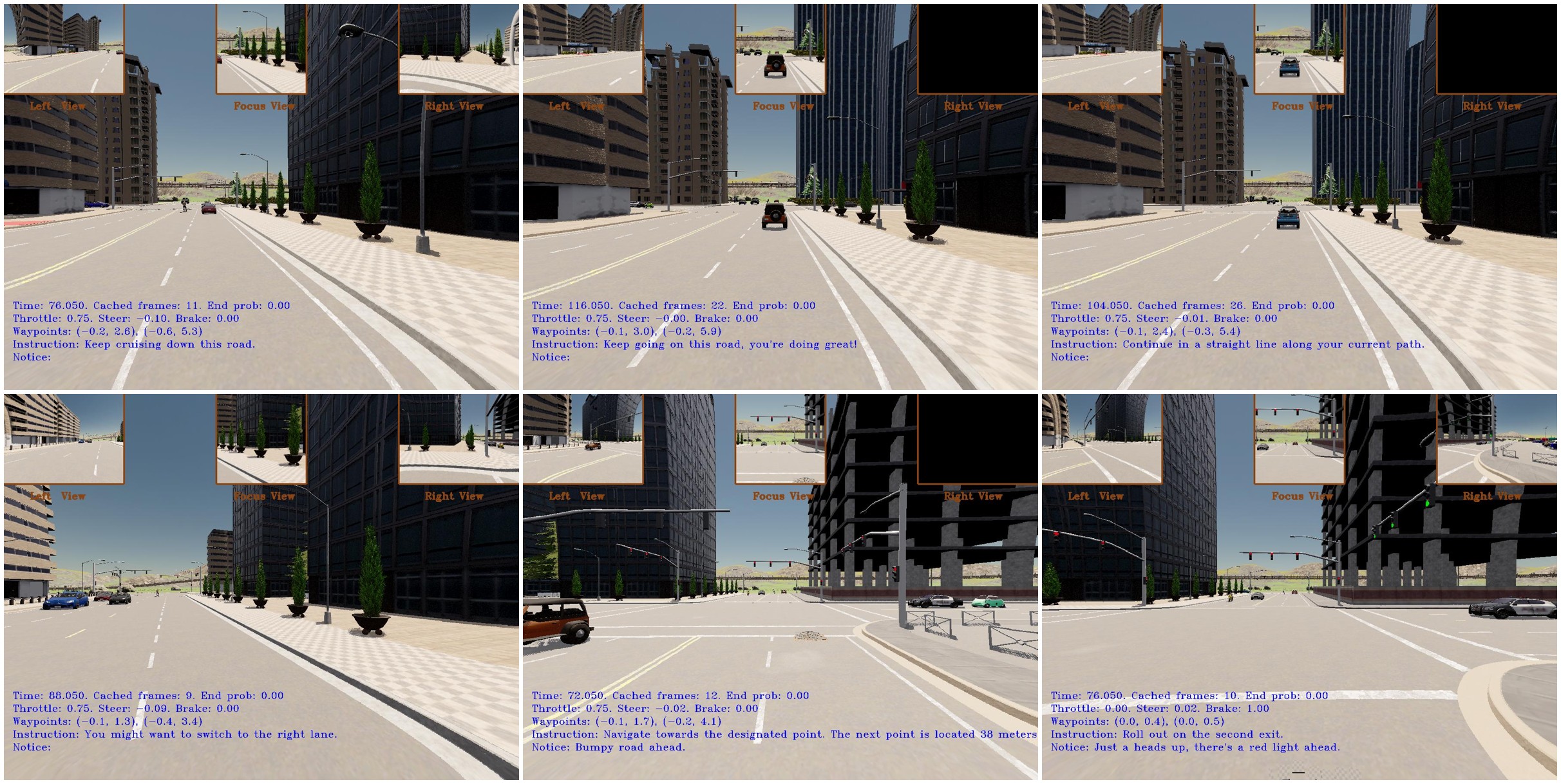}
\caption{Closed-loop driving in town05. Each panel shows the forward view
with side views, the active navigation instruction, the runtime notice,
and the predicted waypoints; the bottom-right panel brakes for a red
light flagged by a notice.}
\label{fig:scenes}
\end{figure}

The frame-skipped middle row of Table~\ref{tab:closedloop} separates the
two ingredients. Running the identical trained expert at the baseline's
10\,Hz replay cadence already reaches a driving score of 34.0, so the
driving-score improvement over the baseline traces to the expert rather
than to control freshness. Freshness contributes elsewhere. Moving from
10\,Hz to 20\,Hz raises completion from 82.1 to 94.0, cuts route
deviations from 11.3 to 4.3 per kilometer, reduces timeouts from 1.3 to
0.08, and lowers red-light violations from 10.4 to 6.9, while the added
exposure costs infraction score (0.45 to 0.37) and leaves the composite
driving score within run-to-run spread. The two rows are not equally
available designs, however. Without cache reuse every step costs
89--169\,ms (Table~\ref{tab:latency}), so frame-skipping is the only form
in which a recompute-based system fits the 50\,ms budget; the cached
32\,ms path is what makes the 20\,Hz row feasible at all.

\subsection{Latency}
\begin{table}[t]
\centering
\caption{Per-stage latency (ms, median, RTX 3090 Ti) at a 40-frame history.
Legacy recomputes Q-Former and backbone over the full history each step;
ours touches the history only through the cache. In parentheses: medians
measured inside the closed loop over 9{,}542 ticks. Totals are measured
end to end rather than summed from stages; the in-loop agent step includes
sensor formatting and harness overhead outside the model.}
\label{tab:latency}
\begin{tabular}{lccc}
\toprule
Stage & Legacy per step & Ours per tick & Ours per $K{=}4$ ticks \\
\midrule
Perception + Q-Former & 26.9 & 24.8 \,(29.2) & --- \\
Backbone full recompute & 61.6 & --- & --- \\
Incremental append & --- & --- & 21.1 \,(10.7) \\
Action expert & --- & 7.5 \,(9.3) & --- \\
\midrule
Total model compute & 88.6 & 32.4 & 5.3 amortized \\
Total agent step, in loop & --- & 58 & --- \\
\bottomrule
\end{tabular}
\end{table}

\begin{figure}[t]
\centering
\includegraphics[width=0.48\linewidth]{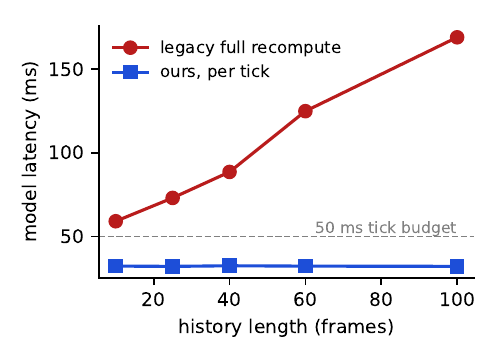}
\hfill
\includegraphics[width=0.48\linewidth]{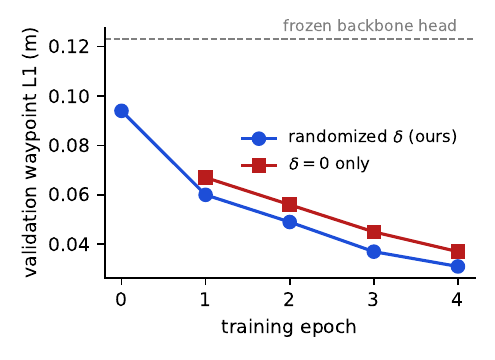}
\caption{Left: per-step model latency versus history length. The legacy
path re-encodes the full history each step and exceeds the 50\,ms tick
budget at every history length, while our per-tick cost stays flat at 32\,ms.
Right: validation waypoint L1 across training epochs; the
randomized-$\delta$ expert leads its $\delta{=}0$ twin at every epoch and
both leave the frozen backbone head far behind.}
\label{fig:curves}
\end{figure}

Figure~\ref{fig:curves} (left) shows the scaling. Legacy cost grows with
history, from 59\,ms at 10 frames to 169\,ms at 100, because every step
re-encodes everything the agent has seen, which is why the
public agent skips ticks. Our per-tick model cost is
flat. The expert reads the cache through attention dominated by its own
ten tokens and measures 7--9\,ms regardless of history length, while cache
maintenance adds an amortized 5\,ms per tick and full rebuilds (11\,ms
median in the loop) occur only at instruction changes, notices, and
episode boundaries. Model compute thus stays approximately 18\,ms below the 50\,ms tick
budget. The measured end-to-end agent step is 58\,ms at the median,
with the difference coming from sensor formatting and evaluation-harness
overhead we did not optimize; in CARLA's synchronous mode the vehicle
still receives fresh control at every tick, and the wall-clock rate is
about 17\,Hz.

\subsection{Zero-Shot Town Transfer}
\label{sec:transfer}
The expert trains on town05 only, so evaluation on town01 and town02,
layouts it has never seen, separates cache-reading skill from town
memorization. The perception encoder, Q-Former, and backbone were
pretrained on every town and are held fixed for both agents; zero-shot
refers strictly to the 337M expert.

\begin{table}[t]
\centering
\caption{Zero-shot transfer to unseen towns, single run per agent and
town. The expert was trained on town05 only.}
\label{tab:transfer}
\begin{tabular}{llcccc}
\toprule
Town & Agent & Control rate & DS $\uparrow$ & RC $\uparrow$ & IS $\uparrow$ \\
\midrule
\multirow{2}{*}{town01 (23 routes)} & LMDrive & 10\,Hz & 20.5 & 40.5 & \textbf{0.56} \\
 & \ours{} & 20\,Hz & \textbf{23.2} & \textbf{84.3} & 0.26 \\
\midrule
\multirow{2}{*}{town02 (15 routes)} & LMDrive & 10\,Hz & 11.6 & 30.7 & \textbf{0.38} \\
 & \ours{} & 20\,Hz & \textbf{28.3} & \textbf{94.4} & 0.29 \\
\bottomrule
\end{tabular}
\end{table}

Route completion transfers almost intact, reaching 94.4 on town02 and 84.3
on town01 against 94.0 in-domain, which indicates that the expert has
learned to read the backbone's cache and the current frame rather than a
particular town's geometry. The advantage over the baseline persists in
both unseen towns and is largest where the baseline degrades most, with
2.4 times its driving score and 3.1 times its completion on town02. Stale
control appears to hurt most where the scene is unfamiliar.

\subsubsection{Long-route transfer probes the boundary.}
We also evaluated eight LangAuto-Long routes in town03, several kilometers
each, a tier absent from the expert's short-route training. Neither agent
attains a usable driving score there. The baseline barely progresses,
completing 12.7\% of route on average for a 5.98 driving score earned
largely by standing still, while ours completes 85.4\% but accumulates
enough vehicle collisions and red-light violations to collapse its penalty
factor to 0.04, for a 2.96 driving score. Completion therefore transfers
even to this tier, while hazard negotiation over long horizons does not.
This points at the training data rather than the architecture, since short
town05 clips contain few dense-traffic signalized encounters; long-route
training clips are the natural next step.

\subsection{Ablations}
\label{sec:ablations}
\subsubsection{Staleness augmentation.}
Randomized-$\delta$ training improves accuracy even when no staleness is
present at test time. Evaluated under the same synchronous protocol
($\delta{=}0$), the randomized expert leads its $\delta{=}0$-only twin at
every epoch (0.060 versus 0.067, then 0.049 versus 0.056, 0.037 versus
0.045, and 0.031 versus 0.037 at convergence), so the augmentation
regularizes even before staleness enters the picture. At deployment the $\delta{=}0$
model additionally faces cache lag it never saw during training on three
of every four ticks.

\subsubsection{Limitations.}
Ours is a simulation study trained on short routes from one town. Hazard
negotiation on long routes does not transfer, per-kilometer vehicle
collisions increase alongside completion, the low-level controller
inherits gains tuned for the slower baseline, and the main comparison
covers two runs per configuration, with single runs for the frame-skip
ablation and the transfer rows and an observed run-to-run spread of up to
1.6 DS. A driving system requires safety
validation far beyond a simulated benchmark, and nothing here should be
read as evidence of road readiness.

\section{Conclusion}
\label{sec:conclusion}

Language models earn their place in a driving stack through understanding,
and an architecture should not also demand reaction speed from the same
component. We decomposed a language-guided driving agent along this line,
pairing a frozen 7B backbone that maintains an incrementally cached
representation of instruction and history with a 337M action expert that
reads this cache through per-layer attention at every simulation tick.
Randomized staleness during training aligns the expert with the
asynchronous cache it meets at deployment, and closed-loop evaluation in
CARLA shows that the two halves of the decomposition contribute
separately: the expert lifts the driving score, and per-tick freshness,
affordable only because the cache holds model cost flat regardless of
history length, lifts route completion to 2.5 times the frame-skipping
baseline. Limitations are cataloged alongside the experiments. The cache interface also opens directions we
have not pursued here, from pretraining the backbone with world models to
enrich what the cache encodes to using fast-slow disagreement as a
training-free trigger for runtime safety supervision.

\bibliographystyle{splncs04}
\bibliography{main}

\end{document}